\documentclass{ceurart}
\usepackage{listings}
\usepackage[inline]{enumitem}
\usepackage{tikz}
\usetikzlibrary{decorations.pathreplacing}
\usepackage{caption}
\usepackage{subcaption}

\begin{document}

\copyrightyear{2024}
\copyrightclause{Copyright for this paper by its authors. Use permitted under Creative Commons License Attribution 4.0 International (CC BY 4.0).}
\conference{BigHPC2024: Special Track on Big Data and High-Performance Computing, co-located with the 3\textsuperscript{rd} Italian Conference on Big Data and Data Science, ITADATA2024, September 17 -- 19, 2024, Pisa, Italy.}
\title{Quantum-Classical Sentiment Analysis}
\author[1]{Mario Bifulco}[email=mario.bifulco@edu.unito.it,url=https://github.com/TheFlonet/qsvm4sentanalysis]
\author[1]{Luca Roversi}[orcid=0000-0002-1871-6109,email=luca.roversi@unito.it,url=https://www.di.unito.it/~rover/]
\address[1]{Università degli studi di Torino, Dipartimento di Informatica, Corso Svizzera 185 - 10149 Torino}

\begin{abstract}
    In this study, we initially investigate the application of a hybrid classical-quantum classifier (HCQC) for sentiment analysis, comparing its performance against the classical CPLEX classifier and the Transformer architecture. Our findings indicate that while the HCQC underperforms relative to the Transformer in terms of classification accuracy, but it requires significantly less time to converge to a reasonably good approximate solution. This experiment also reveals a critical bottleneck in the HCQC, whose architecture is partially undisclosed by the D-Wave property. To address this limitation, we propose a novel algorithm based on the algebraic decomposition of QUBO models, which enhances the time the quantum processing unit can allocate to problem-solving tasks.
\end{abstract}

\maketitle

\section{Introduction}
In natural language processing, the two main challenges in developing new models are the difficulty in acquiring high-quality data and the extensive training times required to make models more expressive\cite{scaling}.

This work focuses on the latter issue. We experiment with unconventional computing architectures, the goal being to assess if and how they can help accelerate training time to obtain more expressive models. To this purpose, our choice is architectures that develop Adiabatic Quantum Computing (AQC), where the technology proposed by D-Wave is considered a standard. The reason is twofold. On one side AQC by its very nature solves minimization problems in the QUBO form (Quadratic Unconstrained Binary Optimization). On the other, the core of many AI problems is minimizing some functions by looking for the values of specific parameters.

We choose SVM\cite{SVM} over more standard Transformers models because preliminary investigations on SVMs that leverage AQC already exist\cite{QSVM} and SVM share some similarities with the attention mechanism of Transformers\cite{TransformerSVM}.

Precalling that, among classification tasks in natural language processing, the binary version of Sentiment Analysis (BSA) aims to separate sentences that convey ``positive'' emotions from those that convey ``negative'' emotions. We reduced the BSA to QUBO and evaluated the following:
\begin{enumerate*}[label=\arabic*)]
    \item performance during classification;
    \item the time required to train the model;
    \item the time required to classify new examples,
\end{enumerate*}
compared to more standard techniques implemented with heuristics and classical architectures. 
To overcome the limited use of the quantum process unit (QPU) by the D-Wave hybrid solver, we also started to investigate algebraic-based alternatives to the proprietary mechanisms that split QUBO problems between QPU and CPU.

\section{Quantum Support Vector Machine for Sentiment Analysis}

We choose TweetEval\cite{TweetEval} to verify the effectiveness of SVM for BSA. TweetEval is considered a standard for comparing different models and contains a sufficiently large and representative number of examples, i.e. Tweets extracted from \url{https://x.com/} and labelled automatically. The ``sentiment'' split of TweetEval includes three classes: positive, negative and neutral. We choose to discard all ``neutral'' samples to avoid introducing errors during learning due to examples belonging to non-expressive classes. Additionally, we normalize the quantity of elements in the positive and negative classes to ensure a balanced dataset.

Since SVMs do not natively support text processing, it is necessary to compute embeddings. Among the various possibilities, SentenceBert\cite{SentenceBert} allows for capturing the contextual information of the entire sentence by producing a single embedding.

For comparison with the classical counterpart, we choose:
\begin{enumerate*}[label=\arabic*)]
    \item the CPLEX\cite{cplex} solver, a widely used optimizer for solving both linear and non-linear programming problems;
    \item RoBERTa\cite{ROBERTA}, a deep-learning model based on BERT\cite{BERT} and the attention mechanism\cite{Attention}.
\end{enumerate*}
RoBERTa allows a fair comparison as the model we use\cite{robertamodel} is fine-tuned on TweetEval.

Below are the results obtained, D-Wave represents our solution.

\paragraph{F1 score} The classification conducted by RoBERTa is significantly better (94.3\%), although the results of both CPLEX and D-Wave are also well above random guesses, respectively with 76.9\% and 76.1\%. 
The slight difference between the solution obtained with CPLEX and that with D-Wave may be due to some limitations of the current hybrid solvers, which restricted the domain of the optimisation variables from real numbers to integers.

\paragraph{Training time} The time required by D-Wave to find an optimal assignment is 60\% less than that of the classical counterpart, with 39.2 seconds against 101.9. 
Although not available, it is reasonable to expect an even better result if compared to the time required by RoBERTa, which we can fairly expect to amount to several hours of training on high-performance machines.

\paragraph{Prediction time} The complexity of RoBERTa architecture also affects the time required for prediction (136.8 seconds) by requiring more computing time than CPLEX or D-Wave, 2.2 and 33.9 seconds respectively. The higher time required is due to D-Wave returning a set of optimal solutions. For each of them, a model is created to apply a majority vote to establish the class in inference. Since no advantages emerge from using several models, it is possible to use only one of the optimal assignments, reducing the time to values comparable to those of CPLEX.

\section{Maximizing quantum boost}

Using the \emph{default} hybrid solver by D-Wave entails an opaque workflow due to the underlying proprietary technology. Examination of the available data, however, shows that to obtain the QPU contributes very marginally, averaging 0.08\%. Since the performance boost should derive from the quantum component, it is worth considering whether it is possible to use a greater amount of it, ideally shifting the entire problem resolution onto the QPU and delegating only pre- and post-processing operations to some CPUs or GPUs.

Direct access to the QPU is possible provided you manually perform certain pre-processing operations of the problem such as:
\begin{enumerate}
    \item Convert the problem into QUBO form i.e:
    \begin{enumerate*}
        \item to incorporate the constraints into the objective as penalty functions via appropriately parameterised Lagrangian relaxation\cite{QbridgeI};
        \item to convert the optimisation variables into binary variables;
        \item to transform the objective function into a minimisation problem.
    \end{enumerate*}
    \item Search for the minor\cite{ME}\cite{MEdwave} associated with the QUBO problem that can be mapped onto the D-Wave QPU's physical graph architecture, Figure \ref{fig:pegasus}.
\end{enumerate}

By performing these steps independently, it is possible to bypass D-Wave's proprietary technology to try to maximise QPU usage. 
The only operation with high computational cost among those described is the search for the minor embedding, which is NP-complete. For this reason, Table \ref{tab:embedding} focuses on the time required to compute the minor embedding graphs representing SVMs. 
We generated the graphs of the SVMs so that all examples were on the x-axis and the number of elements of the positive class was equal to the negative one. This problem does not pose a challenge for classification but allows SVMs of the desired size to be generated easily.

We can see that, for 128 variables, the calculation takes more than 7 minutes and occupies about 38\% of the 5600 qubits available on the QPU. This makes the direct use of the QPU not applicable to real-size problems.

\begin{table}
    \caption{Embedding search time}
    \label{tab:embedding}
    \begin{tabular}{ccccccc}
        \toprule
        Problem nodes & 4 & 8 & 16 & 32 & 64 & 128 \\  
        \midrule
        Embedding Nodes & 4 & 13 & 40 & 138 & 526 & 2117 \\
        Avg Time (s) & 0.2 & 0.3 & 0.6 & 6.1 & 53.4 & 434.2 \\
        \bottomrule
    \end{tabular}
\end{table}

\section{Homebrewing a Hybrid Solver}

Given the practical impossibility of using the QPU directly, it is unavoidable to rely on hybrid solvers. Therefore, we investigate how to design a hybrid solver, called \verb|QSplit|, to increase the use of the QPU, based on a quite simple algebraic technique that decomposes QUBO problems into smaller chunks. Sufficiently small problems can be solved by mapping them on the QPU directly, eventually aggregating the results. This technique is of general application, not limited to QUBO instances we get from SVM.

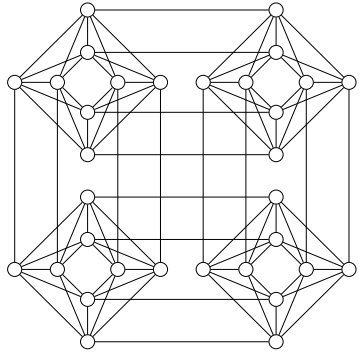
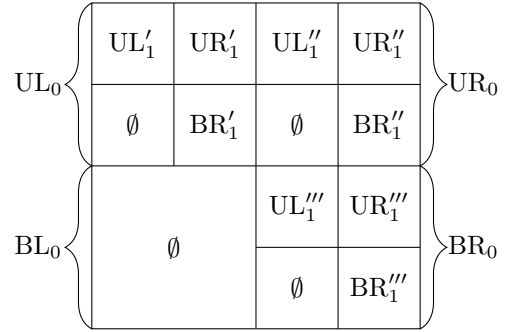
\begin{figure}
    \centering
    \begin{subfigure}[c]{0.54\textwidth}
        \centering
        \resizebox{0.54\textwidth}{!}{%
            \begin{tikzpicture}[main/.style = {draw, circle}] 
                \node[main](1){};
                \node[main](2)[right of=1]{}; 
                \node[main](5)[above right of=2]{};
                \node[main](6)[above of=5]{};
                \node[main](3)[below right of=5]{};
                \node[main](4)[right of=3]{};
                \node[main](7)[below right of=2]{};
                \node[main](8)[below of=7]{};
        
                \draw (1) -- (6);\draw (6) -- (4);\draw (4) -- (8);\draw (8) -- (1);
                \draw (1) -- (5);\draw (5) -- (4);\draw (4) -- (7);\draw (7) -- (1);
                \draw (6) -- (2);\draw (2) -- (8);\draw (8) -- (3);\draw (3) -- (6);
                \draw (2) -- (5);\draw (5) -- (3);\draw (3) -- (7);\draw (7) -- (2);
                \draw (1) -- (2);\draw (5) -- (6);\draw (3) -- (4);\draw (8) -- (7);
        
                \node[main](11)[right of=4]{};
                \node[main](12)[right of=11]{}; 
                \node[main](15)[above right of=12]{};
                \node[main](16)[above of=15]{};
                \node[main](13)[below right of=15]{};
                \node[main](14)[right of=13]{};
                \node[main](17)[below right of=12]{};
                \node[main](18)[below of=17]{};
        
                \draw (11) -- (16);\draw (16) -- (14);\draw (14) -- (18);\draw (18) -- (11);
                \draw (11) -- (15);\draw (15) -- (14);\draw (14) -- (17);\draw (17) -- (11);
                \draw (16) -- (12);\draw (12) -- (18);\draw (18) -- (13);\draw (13) -- (16);
                \draw (12) -- (15);\draw (15) -- (13);\draw (13) -- (17);\draw (17) -- (12);
                \draw (11) -- (12);\draw (15) -- (16);\draw (13) -- (14);\draw (18) -- (17);
        
                \node[main](26)[below of=8]{};
                \node[main](25)[below of=26]{};
                \node[main](22)[below left of=25]{};
                \node[main](21)[left of=22]{};
                \node[main](23)[below right of=25]{};
                \node[main](24)[right of=23]{};
                \node[main](27)[below right of=22]{};
                \node[main](28)[below of=27]{};
        
                \draw (21) -- (26);\draw (26) -- (24);\draw (24) -- (28);\draw (28) -- (21);
                \draw (21) -- (25);\draw (25) -- (24);\draw (24) -- (27);\draw (27) -- (21);
                \draw (26) -- (22);\draw (22) -- (28);\draw (28) -- (23);\draw (23) -- (26);
                \draw (22) -- (25);\draw (25) -- (23);\draw (23) -- (27);\draw (27) -- (22);
                \draw (21) -- (22);\draw (25) -- (26);\draw (23) -- (24);\draw (28) -- (27);
        
                \node[main](36)[below of=18]{};
                \node[main](35)[below of=36]{};
                \node[main](32)[below left of=35]{};
                \node[main](31)[left of=32]{};
                \node[main](33)[below right of=35]{};
                \node[main](34)[right of=33]{};
                \node[main](37)[below right of=32]{};
                \node[main](38)[below of=37]{};
        
                \draw (31) -- (36);\draw (36) -- (34);\draw (34) -- (38);\draw (38) -- (31);
                \draw (31) -- (35);\draw (35) -- (34);\draw (34) -- (37);\draw (37) -- (31);
                \draw (36) -- (32);\draw (32) -- (38);\draw (38) -- (33);\draw (33) -- (36);
                \draw (32) -- (35);\draw (35) -- (33);\draw (33) -- (37);\draw (37) -- (32);
                \draw (31) -- (32);\draw (35) -- (36);\draw (33) -- (34);\draw (38) -- (37);
        
                \draw (6) -- (16);
                \draw (5) -- (15);
                \draw (7) -- (17);
                \draw (8) -- (18);
                \draw (26) -- (36);
                \draw (25) -- (35);
                \draw (27) -- (37);
                \draw (28) -- (38);
                \draw (1) -- (21);
                \draw (2) -- (22);
                \draw (3) -- (23);
                \draw (4) -- (24);
                \draw (11) -- (31);
                \draw (12) -- (32);
                \draw (13) -- (33);
                \draw (14) -- (34);
            \end{tikzpicture} 
        }  
        \caption{Example of 32 qubit QPU\cite{Pegasus}}
        \label{fig:pegasus}
    \end{subfigure}
    \hfill
    \begin{subfigure}[c]{0.45\textwidth}
        \centering
        \begin{tikzpicture}[scale=0.54]
            \draw (0,0) rectangle (8,8);
        
            \draw (0,4) -- (8,4);
            \draw (4,0) -- (4,8);
            \node at (2,2) {$\emptyset$};
            
            \draw (0,6) -- (4,6);
            \draw (2,8) -- (2,4);
            \node at (1,5) {$\emptyset$};
        
            \draw (4,2) -- (8,2);
            \draw (6,4) -- (6,0);
            \node at (5,1) {$\emptyset$};
    
            \draw (4,6) -- (8,6);
            \draw (6,4) -- (6,8);
            \node at (5,5) {$\emptyset$};
        
            \draw[decorate,decoration={brace,amplitude=10pt}] (8,8) -- (8,4) node [black,midway,xshift=20pt] {$\operatorname{UR}_0$};
            \node at (3,7) {$\operatorname{UR}_1'$};
            \node at (7,7) {$\operatorname{UR}_1''$};
            \node at (7,3) {$\operatorname{UR}_1'''$};
    
            \draw[decorate,decoration={brace,amplitude=10pt,mirror}] (0,8) -- (0,4) node [black,midway,xshift=-20pt] {$\operatorname{UL}_0$};
            \node at (1,7) {$\operatorname{UL}_1'$};
            \node at (5,7) {$\operatorname{UL}_1''$};
            \node at (5,3) {$\operatorname{UL}_1'''$};
    
            \draw[decorate,decoration={brace,amplitude=10pt}] (8,4) -- (8,0) node [black,midway,xshift=20pt] {$\operatorname{BR}_0$};
            \node at (3,5) {$\operatorname{BR}_1'$};
            \node at (7,5) {$\operatorname{BR}_1''$};
            \node at (7,1) {$\operatorname{BR}_1'''$};

            \draw[decorate,decoration={brace,amplitude=10pt,mirror}] (0,4) -- (0,0) node [black,midway,xshift=-20pt] {$\operatorname{BL}_0$};
        \end{tikzpicture}
        \caption{Two steps of recursive decomposition}
        \label{fig:qubo}
    \end{subfigure}
    \caption{QPU graph structure (\ref{fig:pegasus}) and decomposition of QUBO matrix (\ref{fig:qubo})}
\end{figure}

Given any QUBO matrix we recursively divide it into four parts as in Figure \ref{fig:qubo}:
\begin{enumerate*}[label=\arabic*)]
    \item $\operatorname{ULs}$ and $\operatorname{BRs}$ are themselves QUBO matrices operating on a partition of the optimization variables;
    \item $\operatorname{UR}$ is not guaranteed to be upper triangular, as required by a QUBO instance, but it retains the information linking the partitions $\operatorname{UL}_0$, $\operatorname{BL}_0$ and $\operatorname{UR}_0$, $\operatorname{BR}_0$ of the variables and we can safely transform it into an upper triangular matrix, namely a QUBO instance;
    \item $\emptyset$ is a matrix composed entirely of zeros, so we ignore it.
\end{enumerate*}

The recursive subdivision can continue until the matrices reach a predetermined size. Compatible with the results of Table \ref{tab:embedding}, directly solving $32 \times 32$ matrices via QPU might be a good compromise for using \verb|QSplit| in real-world contexts.

Let us see how an inductive step of the process works. Let us assume that we have classifications offered by $\operatorname{UL}_1'$, $\operatorname{UR}_1'$, $\operatorname{BR}_1'$. The classification offered by:
\begin{enumerate*}[label=\arabic*)]
    \item $\operatorname{UL}_1'$ and $\operatorname{BR}_1'$ can be combined to generate conflict-free initial assignments called $\operatorname{S_1}$;
    \item $\operatorname{UR}_1'$ offers classifications based on two partitions of variables. In practice, it is solved by incorporating the coefficients of a matrix of zeros of size $\operatorname{UL}_0$. Although this could lead to matrices larger than the bounds tested in Table \ref{tab:embedding}, its graph structure is small enough to allow for manageable computation times directly on the QPU.
\end{enumerate*}
The solutions from $\operatorname{S_1}$ and those obtained from $\operatorname{UR}_1'$ are combined by searching for compatible assignments and marking conflicting variables. From the conflicting values, a QUBO problem is extracted, considering only the rows and columns associated with these variables, which is again solved via QPU. Of course, the proposed method of resolving conflicting assignments could, in the worst case, require reconsideration of the entire problem \(\operatorname{UL}_0\), nullifying the decomposition process and leading to intractable cases. However, the problem did not show up in our tests. From the set of possible assignments obtained by solving conflicting classifications, duplicates are removed and only the $k$ most promising assignments are retained. This is a heuristic that should help avoid an exponential increase in the number of solutions to be aggregated.

\begin{table}
    \centering
    \begin{tabular}{cccc|cc}
        \toprule
        \multicolumn{4}{c}{\texttt{QSplit}} & \multicolumn{2}{c}{\texttt{QPUSampler}} \\
        Cut Dim         & CPU time & QPU time & Solution & Total time & Solution \\
        \midrule
        2               & 417.8    & 0.45     & 0.49     & 303.5      & 0 \\
        4               & 173.3    & 0.35     & 0.41     & 92.1       & 0 \\
        8               & 105.8    & 0.25     & 0.42     & 212.6      & 0 \\
        16              & 49.7     & 0.17     & 0.39     & 119.9      & 0 \\
        32              & 42       & 0.1      & 0.36     & 205.5      & 0 \\
        \bottomrule
    \end{tabular}
    \caption{Results for 128 variables cliques.}
    \label{tab:qsplit}
\end{table}

To test \verb|QSplit| we ran the algorithm on some random problems with 128 variables, collecting:
\begin{itemize}
    \item \texttt{QSplit} information:
    \begin{enumerate*}[label=\arabic*)]
        \item Cut Dim is the size of the problem solved directly via QPU;
        \item CPU time is the total time spent on the CPU;
        \item QPU time is the total time spent on the QPU;
        \item Solution is the best result obtained and normalised with range $[0,1]$.
    \end{enumerate*}
    \item \texttt{QPUSampler} information:
    \begin{enumerate*}[label=\arabic*)]
        \item Total time is the sum of QPU and CPU time;
        \item Solution is the best result obtained and normalised with range $[0,1]$.
    \end{enumerate*}
\end{itemize}

Table \ref{tab:qsplit} shows the results of \texttt{QSplit} compared with the direct execution on the QPU (\texttt{QPUSampler}) for randomly generated problems with 128 variables.

We can see that as the Cut Dim decreases, from 32 to 2, the time spent on the CPU increases significantly. Still, simultaneously, the time spent on the QPU systematically increases, aligning with the objectives of \texttt{QSplit}.
Up to a Cut Dim of 8, the total time required by \texttt{QSplit} is less than that of the direct solution, significantly reducing the time needed for minor embedding computation.

While \texttt{QPUSampler} consistently achieves the optimal value, the solutions proposed by \texttt{QSplit} deteriorate by 35\% to 50\%. It is noteworthy that this performance gap increases as the Cut Dim decreases. This behaviour is reasonable since, by not allowing a holistic view of the problem, \texttt{QSplit} is more likely to encounter local minima that are difficult to bypass.

\section{Conclusion}

Our investigation contributes to giving evidence that quantum computing can bring tangible benefits over traditional methods to solve optimization problems. Our work also confirms that a trade-off between the speed of finding a solution and the quality of the solution is an aspect that must be evaluated on a case-by-case basis.

In many application scenarios, large computational resources are not available, such as in personal computers or embedded systems. In these situations, hybrid solvers such as the one we have presented are good candidates to allow for a good approximation of results while significantly reducing complexity and computational power.

\verb|QSplit| presents a method for handling large QUBO problems that is alternative to those found in the literature\cite{subqubo1}\cite{subqubo2}. It focuses on maximising the use of the QPU. Although this approach is reasonable, it is not guaranteed to lead to an optimal result. The assignment produced can be improved by implementing more refined problem partitioning strategies\cite{bnb}, or by creating work pipelines capable of using a set of methods for finding the optimal assignment\cite{dwavehybrid}.

\bibliography{bibliography.bib}

\begin{thebibliography}{19}
\expandafter\ifx\csname natexlab\endcsname\relax\def\natexlab#1{#1}\fi
\providecommand{\url}[1]{\texttt{#1}}
\providecommand{\href}[2]{#2}
\providecommand{\path}[1]{#1}
\providecommand{\DOIprefix}{doi:}
\providecommand{\ArXivprefix}{arXiv:}
\providecommand{\URLprefix}{URL: }
\providecommand{\Pubmedprefix}{pmid:}
\providecommand{\doi}[1]{\href{http://dx.doi.org/#1}{\path{#1}}}
\providecommand{\Pubmed}[1]{\href{pmid:#1}{\path{#1}}}
\providecommand{\bibinfo}[2]{#2}
\ifx\xfnm\relax \def\xfnm[#1]{\unskip,\space#1}\fi
\bibitem[{Kaplan et~al.(2020)Kaplan, McCandlish, Henighan, Brown, Chess, Child, Gray, Radford, Wu, and Amodei}]{scaling}
\bibinfo{author}{J.~Kaplan}, \bibinfo{author}{S.~McCandlish}, \bibinfo{author}{T.~Henighan}, \bibinfo{author}{T.~B. Brown}, \bibinfo{author}{B.~Chess}, \bibinfo{author}{R.~Child}, \bibinfo{author}{S.~Gray}, \bibinfo{author}{A.~Radford}, \bibinfo{author}{J.~Wu}, \bibinfo{author}{D.~Amodei}, \bibinfo{title}{Scaling laws for neural language models}, \bibinfo{year}{2020}. \href{http://arxiv.org/abs/2001.08361}{{\tt arXiv:2001.08361}}.
\bibitem[{Cortes and Vapnik(1995)}]{SVM}
\bibinfo{author}{C.~Cortes}, \bibinfo{author}{V.~N. Vapnik},
\newblock \bibinfo{title}{Support-vector networks},
\newblock \bibinfo{journal}{Machine Learning} \bibinfo{volume}{20} (\bibinfo{year}{1995}) \bibinfo{pages}{273--297}.
\bibitem[{Willsch et~al.(2019)Willsch, Willsch, Raedt, and Michielsen}]{QSVM}
\bibinfo{author}{D.~Willsch}, \bibinfo{author}{M.~Willsch}, \bibinfo{author}{H.~Raedt}, \bibinfo{author}{K.~Michielsen},
\newblock \bibinfo{title}{Support vector machines on the d-wave quantum annealer},
\newblock \bibinfo{journal}{Computer Physics Communications} \bibinfo{volume}{248} (\bibinfo{year}{2019}) \bibinfo{pages}{107006}.
\bibitem[{Tarzanagh et~al.(2024)Tarzanagh, Li, Thrampoulidis, and Oymak}]{TransformerSVM}
\bibinfo{author}{D.~A. Tarzanagh}, \bibinfo{author}{Y.~Li}, \bibinfo{author}{C.~Thrampoulidis}, \bibinfo{author}{S.~Oymak}, \bibinfo{title}{Transformers as support vector machines}, \bibinfo{year}{2024}. \href{http://arxiv.org/abs/2308.16898}{{\tt arXiv:2308.16898}}.
\bibitem[{Rosenthal et~al.(2017)Rosenthal, Farra, and Nakov}]{TweetEval}
\bibinfo{author}{S.~Rosenthal}, \bibinfo{author}{N.~Farra}, \bibinfo{author}{P.~Nakov},
\newblock \bibinfo{title}{Semeval-2017 task 4: Sentiment analysis in twitter},
\newblock in: \bibinfo{booktitle}{Proceedings of the 11th international workshop on semantic evaluation (SemEval-2017)}, \bibinfo{year}{2017}, pp. \bibinfo{pages}{502--518}.
\bibitem[{Reimers and Gurevych(2019)}]{SentenceBert}
\bibinfo{author}{N.~Reimers}, \bibinfo{author}{I.~Gurevych},
\newblock \bibinfo{title}{Sentence-bert: Sentence embeddings using siamese bert-networks},
\newblock in: \bibinfo{booktitle}{Proceedings of the 2019 Conference on Empirical Methods in Natural Language Processing}, \bibinfo{publisher}{Association for Computational Linguistics}, \bibinfo{year}{2019}.
\bibitem[{IBM(2022)}]{cplex}
\bibinfo{author}{IBM}, \bibinfo{title}{User's manual for cplex}, \bibinfo{year}{2022}. \URLprefix \url{https://www.ibm.com/docs/en/icos/22.1.1?topic=optimizers-users-manual-cplex}.
\bibitem[{Camacho-collados et~al.(2022)Camacho-collados, Rezaee, Riahi, Ushio, Loureiro, Antypas, Boisson, Espinosa~Anke, Liu, Martínez~Cámara et~al.}]{ROBERTA}
\bibinfo{author}{J.~Camacho-collados}, \bibinfo{author}{K.~Rezaee}, \bibinfo{author}{T.~Riahi}, \bibinfo{author}{A.~Ushio}, \bibinfo{author}{D.~Loureiro}, \bibinfo{author}{D.~Antypas}, \bibinfo{author}{J.~Boisson}, \bibinfo{author}{L.~Espinosa~Anke}, \bibinfo{author}{F.~Liu}, \bibinfo{author}{E.~Martínez~Cámara}, et~al.,
\newblock \bibinfo{title}{{T}weet{NLP}: Cutting-edge natural language processing for social media},
\newblock in: \bibinfo{booktitle}{Proceedings of the 2022 Conference on Empirical Methods in Natural Language Processing: System Demonstrations}, \bibinfo{publisher}{Association for Computational Linguistics}, \bibinfo{address}{Abu Dhabi, UAE}, \bibinfo{year}{2022}, pp. \bibinfo{pages}{38--49}.
\bibitem[{Devlin et~al.(2019)Devlin, Chang, Lee, and Toutanova}]{BERT}
\bibinfo{author}{J.~Devlin}, \bibinfo{author}{M.-W. Chang}, \bibinfo{author}{K.~Lee}, \bibinfo{author}{K.~Toutanova}, \bibinfo{title}{Bert: Pre-training of deep bidirectional transformers for language understanding}, \bibinfo{year}{2019}. \href{http://arxiv.org/abs/1810.04805}{{\tt arXiv:1810.04805}}.
\bibitem[{Vaswani et~al.(2017)Vaswani, Shazeer, Parmar, Uszkoreit, Jones, Gomez, Kaiser, and Polosukhin}]{Attention}
\bibinfo{author}{A.~Vaswani}, \bibinfo{author}{N.~Shazeer}, \bibinfo{author}{N.~Parmar}, \bibinfo{author}{J.~Uszkoreit}, \bibinfo{author}{L.~Jones}, \bibinfo{author}{A.~N. Gomez}, \bibinfo{author}{L.~u. Kaiser}, \bibinfo{author}{I.~Polosukhin},
\newblock \bibinfo{title}{Attention is all you need},
\newblock in: \bibinfo{editor}{I.~Guyon}, \bibinfo{editor}{U.~V. Luxburg}, \bibinfo{editor}{S.~Bengio}, \bibinfo{editor}{H.~Wallach}, \bibinfo{editor}{R.~Fergus}, \bibinfo{editor}{S.~Vishwanathan}, \bibinfo{editor}{R.~Garnett} (Eds.), \bibinfo{booktitle}{Advances in Neural Information Processing Systems}, volume~\bibinfo{volume}{30}, \bibinfo{publisher}{Curran Associates, Inc.}, \bibinfo{year}{2017}.
\bibitem[{CardiffNLP(2022)}]{robertamodel}
\bibinfo{author}{CardiffNLP}, \bibinfo{title}{Twitter-roberta-base for sentiment analysis}, \bibinfo{year}{2022}. \URLprefix \url{https://huggingface.co/cardiffnlp/twitter-roberta-base-sentiment-latest}.
\bibitem[{Glover et~al.(2022)Glover, Kochenberger, Hennig, and Du}]{QbridgeI}
\bibinfo{author}{F.~Glover}, \bibinfo{author}{G.~Kochenberger}, \bibinfo{author}{R.~Hennig}, \bibinfo{author}{Y.~Du},
\newblock \bibinfo{title}{Quantum bridge analytics i: a tutorial on formulating and using qubo models},
\newblock \bibinfo{journal}{Annals of Operations Research} \bibinfo{volume}{314} (\bibinfo{year}{2022}) \bibinfo{pages}{141--183}.
\bibitem[{Choi and Vicky(2008)}]{ME}
\bibinfo{author}{Choi}, \bibinfo{author}{Vicky},
\newblock \bibinfo{title}{Minor-embedding in adiabatic quantum computation: I. the parameter setting problem},
\newblock \bibinfo{journal}{Quantum Information Processing} \bibinfo{volume}{7} (\bibinfo{year}{2008}) \bibinfo{pages}{193--209}.
\bibitem[{Cai et~al.(2014)Cai, Macready, and Roy}]{MEdwave}
\bibinfo{author}{J.~Cai}, \bibinfo{author}{W.~G. Macready}, \bibinfo{author}{A.~Roy}, \bibinfo{title}{A practical heuristic for finding graph minors}, \bibinfo{year}{2014}. \href{http://arxiv.org/abs/1406.2741}{{\tt arXiv:1406.2741}}.
\bibitem[{Boothby et~al.(2019)Boothby, Bunyk, Raymond, and Roy}]{Pegasus}
\bibinfo{author}{K.~Boothby}, \bibinfo{author}{P.~Bunyk}, \bibinfo{author}{J.~Raymond}, \bibinfo{author}{A.~Roy}, \bibinfo{title}{Next-Generation Topology of D-Wave Quantum Processors}, \bibinfo{type}{Technical Report}, D-Wave Systems, \bibinfo{year}{2019}.
\bibitem[{Wang et~al.(2012)Wang, L{\"u}, Glover, and Hao}]{subqubo1}
\bibinfo{author}{Y.~Wang}, \bibinfo{author}{Z.~L{\"u}}, \bibinfo{author}{F.~Glover}, \bibinfo{author}{J.-K. Hao},
\newblock \bibinfo{title}{A multilevel algorithm for large unconstrained binary quadratic optimization},
\newblock in: \bibinfo{editor}{N.~Beldiceanu}, \bibinfo{editor}{N.~Jussien}, \bibinfo{editor}{{\'E}.~Pinson} (Eds.), \bibinfo{booktitle}{Integration of AI and OR Techniques in Contraint Programming for Combinatorial Optimzation Problems}, \bibinfo{publisher}{Springer Berlin Heidelberg}, \bibinfo{address}{Berlin, Heidelberg}, \bibinfo{year}{2012}, pp. \bibinfo{pages}{395--408}.
\bibitem[{Albash et~al.(2016)Albash, Spedalieri, Hen, Pudenz, and Tallant}]{subqubo2}
\bibinfo{author}{T.~Albash}, \bibinfo{author}{F.~M. Spedalieri}, \bibinfo{author}{I.~Hen}, \bibinfo{author}{K.~L. Pudenz}, \bibinfo{author}{G.~Tallant},
\newblock \bibinfo{title}{Solving large optimization problems with restricted quantum annealers},
\newblock \bibinfo{year}{2016}.
\bibitem[{Sanavio et~al.(2024)Sanavio, Tignone, and Ercolessi}]{bnb}
\bibinfo{author}{C.~Sanavio}, \bibinfo{author}{E.~Tignone}, \bibinfo{author}{E.~Ercolessi},
\newblock \bibinfo{title}{Hybrid classical–quantum branch-and-bound algorithm for solving integer linear problems},
\newblock \bibinfo{journal}{Entropy} \bibinfo{volume}{26} (\bibinfo{year}{2024}) \bibinfo{pages}{345}.
\bibitem[{Booth et~al.(2017)Booth, Reinhardt, and Roy}]{dwavehybrid}
\bibinfo{author}{M.~Booth}, \bibinfo{author}{S.~P. Reinhardt}, \bibinfo{author}{A.~Roy}, \bibinfo{title}{Partitioning Optimization Problems for Hybrid Classical/Quantum Execution}, \bibinfo{type}{Technical Report}, D-Wave Systems, \bibinfo{year}{2017}.

\end{thebibliography}

\end{document}